\documentclass[letterpaper, 10 pt, conference]{ieeeconf}  

\IEEEoverridecommandlockouts                              

\usepackage{graphics} 
\usepackage{epsfig} 
\usepackage{times} 
\usepackage{amsmath} 
\usepackage{amssymb}  
\usepackage[bookmarks=true]{hyperref}
\usepackage{xcolor}
\usepackage{color}
\usepackage[ruled,vlined]{algorithm2e}

\definecolor{somegray}{rgb}{0.5, 0.5, 0.5}
\newcommand{\darkgrayed}[1]{\textcolor{somegray}{#1}}
\makeatletter
\newcommand*\titleheader[1]{\gdef\@titleheader{#1}}
\AtBeginDocument{%
  \let\st@red@title\@title
  \def\@title{%
    \vskip-3em
    \bgroup\normalfont\large\centering\@titleheader\par\egroup
    \vskip1.5em\st@red@title}
}
\makeatother

\titleheader{ \darkgrayed{This paper has been accepted for publication at the
IEEE/RSJ International Conference on Intelligent \\ Robots and Systems (IROS), Las Vegas, 2020. 
\copyright IEEE} }

\title{\LARGE \bf
	Learning High-Level Policies for Model Predictive Control
}

\author{
    Yunlong Song,
    Davide Scaramuzza 
    \thanks{    The authors are with the Robotics and Perception Group, Dep.  of Informatics, University of Zurich, and Dep. of Neuroinformatics, University of Zurich and ETH Zurich,  Switzerland. {\tt\small http://rpg.ifi.uzh.ch}.
    This research was supported by the National Centre of Competence in 
    Research (NCCR) Robotics through the Swiss National Science Foundation, 
    the SNSF-ERC Starting Grant, and the European Union’s Horizon 2020 Research
    and Innovation program through the AERIAL-CORE project (H2020-2019-871479).
    }
}

\newcommand{\bs}{\boldsymbol}
\newcommand{\mb}{\mathbf}

\begin{document}

\maketitle
\thispagestyle{empty}
\pagestyle{empty}

\begin{abstract}
The combination of policy search and deep neural networks holds
the promise of automating a variety of decision-making tasks. 
Model Predictive Control~(MPC) provides robust solutions to robot control tasks
by making use of a dynamical model of the system and solving 
an optimization problem online over a short planning horizon.
In this work, we leverage probabilistic decision-making approaches and the generalization capability of 
artificial neural networks to the powerful online optimization by learning a deep high-level policy for the MPC~(High-MPC).
Conditioning on robot's local observations, the trained neural network policy is 
capable of adaptively selecting high-level decision variables for the low-level MPC controller, 
which then generates optimal control commands for the robot.
First, we formulate the search of high-level decision variables for MPC as 
a policy search problem, specifically, a probabilistic inference problem.
The problem can be solved in a closed-form solution.
Second, we propose a self-supervised learning algorithm for learning a neural 
network high-level policy, which is useful for online hyperparameter adaptations in highly dynamic environments.  
We demonstrate the importance of incorporating the online adaption into autonomous robots
by using the proposed method to solve a challenging control problem, 
where the task is to control a simulated quadrotor to fly through a swinging gate.
We show that our approach can handle situations that are difficult 
for standard MPC. 
\end{abstract}

~\\
Code:
\url{https://github.com/uzh-rpg/high_mpc}

\section{Introduction}
    \label{section: intro}
    Model Predictive Control~(MPC)~\cite{rawlings2009model, mayne2014model} is a powerful
    approach for dealing with complex systems with the capability of handling multiple inputs and outputs.
    MPC has become increasingly popular for robot control due to 
    its robustness to model errors and its capability of incorporating actions limits
    and solving optimizations online.
    However, many popular MPC algorithms~\cite{PAMPC, rawlings2009model, kamel2017linear} rely on tools from constrained optimization, 
    which means that convexification, such as a quadratic formulation of the cost function, 
    and approximations of the dynamics are required~\cite{infompc}.
    The requirement of solving constrained optimization online limits the usage of MPC
    for dealing with high-dimensional states and complex cost formulation.
    
    Model-free Reinforcement Learning~(RL) offers the promise of automatically learning
    hard-to-engineer policies for complex tasks~\cite{kober2009policy, rwr, deisenroth2013survey}.
    In particular, in combination with deep neural networks, 
    deep RL~\cite{mnih2015human, rl_quad, sac} optimizes policies that are capable of mapping
    high-dimensional sensory inputs directly to control commands.
    However, the learning of deep neural network policies is highly 
    data-inefficient and suffers from poor generalization. 
    In addition, these methods typically provide little safety or stability guarantees for 
    the system, which is particularly problematic when working with physical robots.
    
    Instead of learning end-to-end control policies that map observations directly to 
    robot's control commands, we consider the problem of learning a high-level policy,
    where the policy chooses task-dependent decision variables for a low-level MPC controller.
    The MPC takes the decision variables as inputs and generates optimal control commands 
    that are eventually executed on the robot.
    The policy parameters we are trying to learn can be hyperparameters that are hard-to-identify by human experts 
    or a compact representation of high-dimensional states (see Section\ref{section: method}).
    
    \begin{figure}[t]
     \centering
     \includegraphics[width=0.4\textwidth]{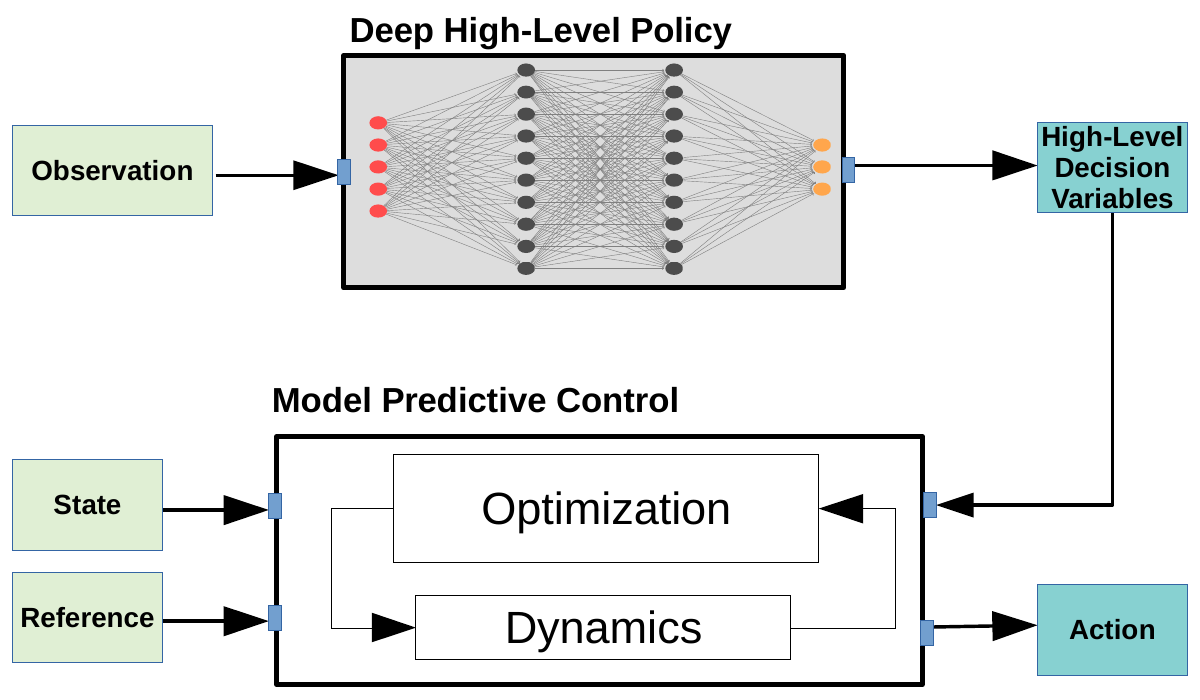}
     \caption{An overview of our approach for online adaptations of model predictive control using a learned deep high-level policy. 
        The neural network policy is trained using self-supervised learning~(Algorithm \ref{algo: online_mpc}).}
     \label{fig: method_overview}
    \end{figure}
    
    \textbf{Contributions:}
    In this work, we leverage intelligent decision-making approaches to 
    the powerful model predictive control.
    First, we formulate the search of high-level decision variables for MPC
    as a probabilistic policy search problem.
    We make use of a weighted maximum likelihood approach~\cite{rwr} for learning the policy parameters, 
    since it allows a closed-form solution for the policy update.
    Second, we propose a novel self-supervised learning algorithm for learning a neural 
    network high-level policy.
    Conditioning on the robot's observation in a rapidly changing environment, 
    the trained policy is capable of adaptively selecting decision variables for MPC. 
    We demonstrate the effectiveness of our approach, which incorporates a learned High-level policy into a MPC (High-MPC), 
    by solving a challenging task of controlling a quadrotor to fly through a fast swinging gate.

\section{Related Work}
    The study of combining machine learning or reinforcement learning with model predictive control 
    has been conducted in learning-based control.
    
    \textbf{Sampling-based MPC} are discussed in~\cite{infompc, mppi},
    in which the MPC optimizations are capable of handling complex cost criteria and 
    making use of learned neural networks for dynamics modelling.
    A crucial requirement for the sampling-based MPC is to generate a large number of samples 
    in real time, where the sampling is generally performed in parallel using graphics processing units~(GPUs).
    Hence, it is computationally expensive to run sampling-based MPC in real time.
    These methods generally focus on learning dynamics for tasks where a dynamical model of the robots or its environment is 
    difficult to derive analytically, such as aggressive autonomous driving around a dirt track~\cite{infompc}.
    
    \textbf{MPC-guided policy search}~\cite{guidedMPC, quad_mpc_guided, levine2016end} are methods that study the problems
    of learning a deep neural network control policy using an MPC as the teacher, and hence, they transform policy 
    search into a supervised learning fashion.
    The trained end-to-end control policy can forgo the need for explicit state estimation and directly map sensor observations to actions.
    MPC-guided policy search has been demonstrated to be more data efficient than standard model-free 
    reinforcement learning. However, it suffers from the problem of poor generalizations and stability.
    
    \textbf{Supervised learning for MPC}~\cite{elia_beauty, cnn_mpc, kahn2020badgr, deep_drone} has been studied in the literature. 
    In~\cite{elia_beauty, deep_drone}, the authors proposed to combine a CNN-based high-level policy 
    with a low-level MPC controller to solve the problem of navigating a quadrotor to pass through multiple gates.
    The trained policy predicts three-dimensional poses of the gate's center from image observations, and then, 
    the MPC outputs control commands for the quadrotor such that it navigates to the predicted waypoints.
    Similarly, the method in \cite{cnn_mpc} tackles an aggressive high-speed autonomous driving problem by 
    using a CNN-based policy to predict a cost map of the track, which is then directly used for online 
    trajectory optimization. Here, the deep neural network policies are trained using supervised learning, 
    which requires ground-truth labels.


\section{Background} 
\label{section: background}

\subsection{Model Predictive Control}
    We consider the problem of controlling an nonlinear deterministic dynamical system 
    whose dynamics is defined by a differential equation $\mb{\dot{x}}_t = f(\mb{x}_t, \mb{u}_t)$,
    where $\mb{x}_t \in \mathbb{R}^{n}$ is the state vector, $\mb{u}_t \in \mathbb{R}^{m}$ is 
    a vector of the control command, and $\mb{\dot{x}}_t \in \mathbb{R}^{n}$ is the derivative of current state.
    In model predictive control, we approximate the actual continuous time differential
    equation using a set of discrete time integration $\mb{x}_{h+1}= \mb{x}_h + d_t * \hat{f}(\mb{x}_h, \mb{u}_h)$,
    with $d_t$ as the time interval between consecutive 
    states and $\hat{f}$ as an approximated dynamical model. 
    
    At every time step $t$, the system is in state $\mb{x}_t$. MPC takes the current state $\mb{x}_t$ 
    and a vector of additional references $\mb{p}$ as input. 
    MPC produces a sequence of optimal system states and control commands
    $\bs{\tau} = \{ (\mb{x}_{1}, \mb{u}_{1}),  \cdots, (\mb{x}_{H-1}, \mb{u}_{H-1}), \mb{x}_{H}\}$
    by solving an optimization online, using a mulitple-shooting scheme.
    The first control command is applied to the system, after which the optimization problem
    is solved again in the next state.
    MPC requires minimizing a quadratic cost over a fixed time 
    horizon $H$ at each control time step by solving a constrained optimization:
    \begin{equation}
        \label{rq: mpc_obj}
        \begin{aligned}
         &\min_{ \mb{u}_{1:H}, \mb{x}_{1:H} } & J = \sum_{h=1}^{H} c(\mb{x}_h, \mb{u}_h, \mb{p}, \mb{z}) \\
         &\text{subject to} & \quad \mb{g}(\mb{x}, \mb{u}) = 0, \quad \mb{h}(\mb{x}, \mb{u}) \leq 0  \\
         && \mb{x}_{h+1} = \mb{x}_h + d_t * \hat{f}(\mb{x}_h, \mb{u}_h), & \quad
         \mb{x}_1 = \mb{x}_\text{init}
        \end{aligned}
    \end{equation}
    where $\mb{g}(\mb{x}, \mb{u})$ represents equality constraints and
    $\mb{h}(\mb{x}, \mb{u})$ represents inequality constraints. 
    Here, $\mb{p}$ is a vector of reference states that are normally determined
    by a path planner and are directly related to the task goal.
    We represent a vector of high-level variables as $\mb{z}$, which has to be defined in
    advance by human experts, or learned using our policy search algorithm~(Sec.~\ref{section: method}).
    
\subsection{Episode-based Policy Search}
\begin{figure}[!htp]
         \centering
         \includegraphics[width=0.3\textwidth]{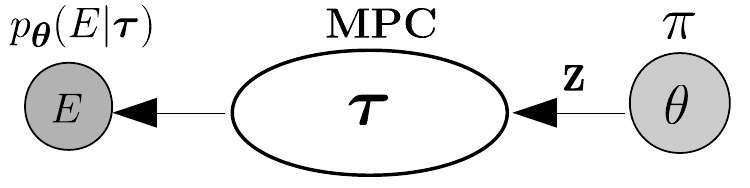}
         \caption{Graphical model for learning a high-level policy~$\pi_{\mb{\theta}}$ for MPC.}
         \label{fig: vi}
    \end{figure}

    We summarize episode-based policy search by following the derivation from~\cite{deisenroth2013survey}.
    Unlike step-based policy search~\cite{sac, Sutton1998},
    which explores in the action space by adding exploration noise directly to the executed actions,
    episode-based policy search perturbs the parameters of a low-level controller in parameter space~\cite{deisenroth2013survey}.
    This kind of exploration is normally added in the beginning of an episode and a reward function
    is used to evaluate the quality of trajectories that are generated by sampled parameters.
    A list of episode-based policy search algorithms have been discussed in literature~\cite{rwr, cem, h_reps, deisenroth2013survey}. 
    We focus on a probabilistic model in which the search of high-level parameters for the low-level controller 
    is treated as a probabilistic inference problem. 
    A visualization of the inference problem is given in Fig~\ref{fig: vi},
    the graphical model is inspired by~\cite{deisenroth2013survey}.
    
    We make use of an MPC as the low-level controller where the decision variables in MPC is represented 
    as a vector of unknown variables $\mb{z}$.
    We define a reward function as $R(\bs{\tau})$, which is used to evaluate
    the goodness of the MPC solution $\bs{\tau}$ with respect to the given task.
    The goal of policy search is to find the optimal policy $\pi(\bs{\theta}^{\ast})$ such that 
    it automatically selects the high-level variables $\mb{z}$ for the MPC. 
    Therefore, it is equivalent to maximize an expectation of the reward signal.
    Here, the reward function is different from the cost function optimized by the MPC, 
    but directly related to the task goal.
    
    To formulate the policy search as a latent variable inference problem, similar to~\cite{deisenroth2013survey},
    we introduce a binary ``reward event” as the observation, denoted as $E = 1$.
    Maximizing the reward signal implies maximizing the probability of this ``reward event”. 
    This leads to the following maximum likelihood problem~\cite{deisenroth2013survey}:
    \begin{equation}\label{eq: max_log_pro}
        \max_{\bs{\theta}} \quad \log p_{\bs{\theta}}(E=1) = \log \int_{\bs{\tau}} p(E|\bs{\tau}) p_{\bs{\theta}} (\bs{\tau)} d \bs{\tau},
    \end{equation}
    which can be solved efficiently using Monte-Carlo Expectation-Maximization~(MC-EM)~\cite{kober2009policy, vlassis2009model}. 
    MC-EM algorithms find the maximum likelihood solution for
    the log marginal-likelihood~(\ref{eq: max_log_pro}) by introducing a variational distribution~$q(\bs{\tau})$,
    and then, decompose the marginal log-likelihood into two terms:
    \begin{equation}
        \log p_{\bs{\theta}}(E=1) = \mathcal{L}_{\bs{\theta}} (q(\bs{\tau}) ) + \text{KL}(q(\bs{\tau}) || p_{\bs{\theta}}(\bs{\tau} | E))
    \end{equation}
    where $\mathcal{L}_{\bs{\theta}} (q(\bs{\tau}))$ is the lower bound of~$\log p_{\bs{\theta}}(Z=1)$. 
    
    The MC-EM algorithm is an iterative method alternates between performing an Expectation~(E) step and a Maximization~(M) step.
    In the expectation step, we minimize the Kullback–Leibler~(KL) divergence~$\text{KL}(q(\bs{\tau}) || p_{\bs{\theta}}(\bs{\tau} | E))$,
    which is equivalent to setting $q(\bs{\tau})=p_{\bs{\theta}}(\bs{\tau} | E) \propto p(E|\bs{\tau})p_{\bs{\theta}}(\bs{\tau})$.
    In the maximization, we use the sampled distributions for estimating the complete-data log-likelihood 
    by maximizing the following weighted maximum likelihood objective:
    \begin{equation}
        \label{eq: wml}
        \bs{\theta}^{\ast} = \arg \max_{\bs{\theta}} \left\{ \sum_i d^{[i]} \log \pi (\bs{z}^{[i]}; \bs{\theta}) \right\}
    \end{equation}
    where $d^{[i]}=p(E|\bs{\tau}^{[i]})$ is an improper probability distribution for the trajectory $\bs{\tau}^{[i]}$.
    The trajectory $\bs{\tau}^{[i]}$ is collected by solving an MPC optimization problem using $\mb{z}^{[i]}$. 
    The solution for updating the policy parameters $\bs{\theta}$ has a closed-form expression.

\section{Methodology}
\label{section: method}
\subsection{Problem Formulation}
    We make use of a Gaussian distribution $\pi_{\bs{\theta}} \sim \mathcal{N}(\bs{\mu}, \bs{\Sigma})$ 
    to model the high-level policy, where $\bs{\mu}$ is the mean vector,
    $\bs{\Sigma}$ is a covariance matrix, and hence, $\bs{\theta}=\left\{\bs{\mu}, \bs{\Sigma}\right\}$
    represents all policy parameters.
    We design a model predictive control with a vector of unknown decision variables~$\bs{z}$ to be specified.
    The variables are directly related to the goal of a task and have to be specified in advance 
    before MPC solves the optimization problem. 
    MPC produces a trajectory $\bs{\tau}$ that consists of a sequence
    of optimal system states and control commands $(\mb{s}, \mb{u})$.
    The cost function is defined by the variables and additional references states, 
    such as a target position or a planned trajectory. 
    
    We define a reward function~$R(\bs{\tau})$ which evaluates the 
    goodness of the predicted trajectory~$\bs{\tau}$ with respect to the task goal.
    The design of this reward function is more flexible than the cost function optimized by MPC, 
    which allows us to work with complex reward criteria, such as exponential reward, 
    discrete reward, and even sparse reward.
    For example, we can compute the reward by counting the total number of 
    non-collision states in the predicted trajectory. 
    Maximizing this reward can hence find the optimal collision free trajectory. 

\subsection{Probabilistic Policy Search for MPC} \label{subsection: learn_up}
    We first focus on solving the problem of learning a high-level policy $\pi_{\bs{\theta}}$ that does
    not depend on robot's observations, where our goal is to find an optimal policy which maximizes the expected reward
	of predicted trajectories denoted as $\bs{\tau}$.
	We used a weighted maximum likelihood algorithm to 
	solve the maximum likelihood estimation problem, where maximizing the reward is 
	equivalent to maximizing the probability of the binary ``event", 
	denoted as $p_{\bs{\theta}}(E|\bs{\tau})$ (Section~\ref{section: background}). 
    
    The maximization problem corresponds to weighted maximum likelihood estimation of $\pi_{\bs{\theta}}$
    where each sample $\bs{\tau}^{[i]}$ is weighted by $d^{[i]}=p(E|\bs{\tau})$.
    To transform the reward signal $R(\bs{\tau}^{[i]})$ of a sampled trajectory $\bs{\tau}^{[i]}$
    into a probability distribution $d^{[i]}$, we use the exponential transformation~\cite{deisenroth2013survey}:
    \begin{equation}
        \label{eq: d}
        d^{[i]} = \exp{ \left\{ \beta R(\bs{\tau}^{[i]}) \right\} }
    \end{equation}
    where the parameter $\beta \in \mathbb{R}_{+}$ denotes the inverse temperature of the soft-max distribution,
    higher value of $\beta$ implies more greedy policy update. 
    A comparison of using different $\beta$ for the policy update is shown in Fig.~\ref{fig: learning_progress}.
    A complete episode-based policy search for learning a high-level policy in MPC 
    is given in Algorithm~\ref{algo: auto_mpc}.
            \begin{algorithm}[!htp]
    \label{algo: auto_mpc}
    \caption{{\bf Probabilistic Policy Search for MPC} \label{algo: ep_mpc}}
	\KwIn{ $\pi_{\bs{\theta}}(\bs{\mu}, \bs{\Sigma}), N, \text{MPC}, \mb{x}_0, \mb{p}$}
	\textbf{While not converged} \\
	\quad Sample variables: $\bs{z}^{[i]} \sim \pi_{\bs{\theta}}(\bs{z} | \bs{\mu}, \bs{\Sigma})_{i=1...N}$\\
	\quad Sample trajectories: $\bs{\tau}^{[i]} = \text{MPC.solve}(\mb{x}_0, \bs{z}^{[i]}, \mb{p})$ \\
	\quad \textbf{Expectation:} \\
	\quad \quad $d^{[i]} = \exp{ \left\{ \beta R(\bs{\tau}^{[i]}) \right\}}$ \\
	\quad \textbf{Maximization:} \\
	\quad \quad $\bs{\mu}=\left(\sum_{i=1}^N d^{[i]} \mb{z}^{[i]}\right)/\left(\sum_{i=1}^N d^{[i]}\right)$ \\
	\quad \quad $\bs{\Sigma}=\left( \sum_{i=1}^N d^{[i]}(\mb{z}^{[i]}-\bs{\mu})(\mb{z}^{[i]}-\bs{\mu})^T\right)/Y$ \\
	\quad \quad $Y= \left( (\sum_{i=1}^{N} d^{[i]})^2 - \sum_{i=1}^N (d^{[i]})^2 \right) / \left(\sum_{i=1}^N d^{[i]} \right)$ \\
	$\rightarrow \bs{\theta}_\text{new} = [\bs{\mu}, \bs{\Sigma}]$\\
	\KwOut{Learned high-level policy $\pi_{\bs{\theta}}(\bs{\mu}^{\ast}, \bs{\Sigma}^{\ast})$}
\end{algorithm}

    We represent our policy $\pi_{\bs{\theta}}$ using a normal distribution
    with randomly initialized policy parameters $\bs{\theta}$.
    We consider the robot at a fixed state $\mb{x}_0$, which does not change during the learning.
    At the beginning of each training iteration, we randomly sample a list of parameters of length $N$ 
    from the current policy distribution~$\pi_{\bs{\theta}}$ and evaluate the parameters via a predefined
    reward function $R(\bs{\tau})$, where $\bs{\tau}^{[i]}$ are the trajectories predicted by solving the MPC
    with sampled variables $\bs{z}^{[i]}$.
    
    In the Expectation step, we transform the computed reward signal $R(\bs{\tau})$ into
    a non-negative weight $d^{[i]}$~(improper probability distribution) via the exponential transformation~(\ref{eq: d}). 
    In the Maximization step, we update the policy parameters by optimizing the 
    weighted maximum likelihood objective~$(\ref{eq: wml})$, 
    where the policy parameter, both the mean and the covariance, are updated using a closed-form expression.
    We repeat this process until the expectation of sampled reward converges. 
    Here, $\mb{p}$ is a vector of auxiliary variables.
	After training (during policy evaluation), we simply take the mean vector of the Gaussian policy 
	as the optimal decision variables for the MPC.
	Therefore, $\mb{z}=\bs{\mu}^{\ast}$ is the optimal MPC decision variables found by our approach.

\subsection{Learning A Deep High-Level Policy}
    We extend Algorithm~\ref{algo: auto_mpc} of learning a high-level policy to learning a deep neural
    network high-level policy, where the trained neural network policy is capable of 
    selecting adaptive decision variables for the MPC given different observations of the robot.
    Such properties are potentially useful for the robot to adapt its behavior online in 
    a highly dynamic environment.
    For example, it is important to use an adaptive control scheme for mobile robots since 
    the robot's dynamics and its surrounding environment changes frequently.
    
    First, we characterize an observation vector of the robot as $\mb{o}$, where the observation can be either
    high-dimensional sensory inputs, such as images, or low-dimensional states, such as the robot's pose. 
    Second, we define a general-purpose neural network denoted as $f_{\mb{\Phi}}$, with $\mb{\Phi}$ being the 
    network weights to be optimized. 
    We train the deep neural network policy by combining the episode-based policy search~(Algorithm~\ref{algo: auto_mpc})
    with a self-supervised learning approach. 
    Our algorithm of learning a deep high-level policy is summarized in Algorithm~\ref{algo: online_mpc}.
    \begin{algorithm}[!htp]
    \caption{{\bf Learning A Deep High-Level Policy}   \label{algo: online_mpc}}
    \KwIn{ $f_{\mb{\Phi}}, \mathcal{D}=\{ \},$ \textbf{Algorithm}~\ref{algo: auto_mpc} }
	\textbf{Data collection (repeat)} \\
	\quad  Randomly reset the system: $\mb{x}_t, \mb{o}_t, \mb{p}_t, t=0$\\
	\quad  While not done: \\
	\quad \quad $(\mb{z}_t = \bs{\mu}^{\ast}) \leftarrow$ \textbf{Algorithm~\ref{algo: auto_mpc}} ($\mb{x}_0=\mb{x}_t, \mb{p}_t$) \\
	\quad \quad Data collection: $\mathcal{D} \leftarrow \mathcal{D} \cup \left\{\mb{o}_t, \mb{z}_t\right\}$ \\
	\quad \quad MPC optimization: $\mb{u}^{\ast}_t = \text{MPC.solve}(\mb{x}_t, \mb{z}_t, \mb{p}_t)$\\
	\quad \quad System transition: $\mb{x}_{t} \leftarrow f(\mb{x}_t, \mb{u}^{\ast}_t)$ \\
	\textbf{Policy learning} \\
	\quad $\mb{\Phi}_\text{new} = \arg \min_{\mb{\Phi}} \| f_{\mb{\Phi}}(\mb{o}_t) - \mb{z}_t \|^2$ \\
	\KwOut{Learned deep high-level policy $f_{\mb{\Phi}^{\ast}}$}
\end{algorithm}
    
    We divide the learning process into two stages: 1) data collection, 2) policy learning.
    In the data collection stage, we randomly initialize the robot in a state $\mb{x}_t$ and find 
    the optimal decision variables $\mb{z}^{\ast}_t$ via Algorithm~\ref{algo: auto_mpc}.
    We aggregate our dataset by $\mathcal{D} \leftarrow \mathcal{D} \cup (\mb{o}_t, \mb{z}^{\ast}_t)$,
    where $\mb{o}_t$ is the current observation of the robot.
    An sequence of optimal control actions $\mb{u}^{\ast}_t$ are computed by solving the MPC optimization, 
    given the current state $\mb{x}_t$ of the robot and the learned variable $\mb{z}^{\ast}_t$.
    The first control command is applied to the system, subsequently, the robot transitions to the next state.
    Incrementally, we collect a set of data that consists of a variety of observation-optimal-variables pairs $(\mb{o}_t, \mb{z}^{\ast}_t)$. 
    In the policy learning stage, we optimize the neural network by minimizing the mean-squared-error
    between the labels $\mb{z}^{\ast}_t$ and the prediction of the network $f_{\mb{\Phi}}(\mb{o}_t)$,
    using stochastic gradient descent.

\section{Experiments}
\label{section: experiments}

\subsection{Problem Formulation}
    \subsubsection{Passing Through a Fast Moving Gate}
    To demonstrate the effectiveness of our approach, we aim at solving a challenging control problem. Our task
    is to maneuver a quadrotor to pass through the center of a swinging gate that hangs from the ceiling via a cable. 
    We assume that the gate oscillates in a same two-dimensional plane (Fig.~\ref{fig: upmpc_examples}).
    Thus, we model the motion of the gate as a simple pendulum.
    Such a quadrotor control problem can be solved via a traditional modular planning-tracking pipeline, where
    an explicit trajectory generator, such as a minimum snap trajectory~\cite{minimum_snap} or motion primitives~\cite{effi_quad}
    is combined with a low-level controller.
    To forgo the need for an explicit trajectory generator, 
    we intend to solve this problem using our proposed High-MPC, where we 
    make use of a high-level policy to adaptively select a decision variable for a 
    low-level MPC controller. 
    Our approach automatically find an optimal trajectory for flying through the gate by
    solving an adaptive MPC optimization online, 

    \textbf{Quadrotor Dynamics:}
    We model the quadrotor as a rigid body controlled by four motors.
    We use the quadrotor dynamics proposed in~\cite{effi_quad}:
    \begin{align*}
    	\mathbf{\dot{p}}_{WB} &= \mathbf{v}_{WB} \\
    	\mathbf{\dot{v}}_{WB} &= \mathbf{q}_{WB} \odot \mathbf{c} - \mathbf{g}\\
    	\mathbf{\dot{q}}_{WB} &= \frac{1}{2} \mathbf{\Lambda} ( \boldsymbol{\omega}_{B}) \cdot \mathbf{q}_{WB}
    \end{align*}
    where $\mathbf{p}_{WB}=[x_q, y_q, z_q]^T$ and $\mathbf{v}_{WB}=[v_{q, x},v_{q, y},v_{q, z}]^{T}$ are the position and velocity of the quadrotor
    in the world frame~${W}$.
    We use a unit quaternion $\mathbf{q}_{WB}=[q_{q, w},q_{q, x},q_{q, y},q_{q, z}]^{T}$ to represent the orientation of the quadrotor
    and use $\bs{\omega}_{B}= [\omega_x, \omega_y, \omega_z]^T$ to denote the body rates 
    (roll, pitch, and yaw respectively) in the body frame~${B}$.
    Here, $\mathbf{g}=[0, 0, -g_z]^{T}$ with $g_z=9.81 m/s^2$ is the gravity vector,
    and $\mathbf{\Lambda} (\bs{\omega}_{B})$ is a skew-symmetric matrix.
    Finally, $\mb{c}=[0, 0, c]^T$ is the mass-normalized thrust vector. 
    We use a state vector $\mb{x}_q=[\mb{p}_{WB}, \mb{v}_{WB}, \mb{q}_{WB}]$
    and an action vector $\mb{u}_q = [c, \omega_x, \omega_y, \omega_z]$ to denote the
    quadrotor's states and control commands separately.
    
    \textbf{Pendulum Dynamics:}
    We use a simple pendulum which is modeled as a bob of mass $m_p$ attached to the end of a massless cord $L$. 
    The cord is hinged at a fixed pivot point denoted as $\bs{P}_{WP}=[x_f, y_f, z_f]$.
    The pendulum is subject to three forces: the gravity, the tension force exerted by the cord upon the bob, 
    and a damping force due to friction and air drag. 
    The damping force is proportional to the angular velocity $\dot{\theta}$ and denoted as $f_d=-b * \dot{\theta}$, 
    where $b \in \mathbb{R}_{+}$ is a damping factor. 
    Hence, we use the following dynamical model
    \begin{align*}
        \dot{\theta} & = \theta_v \\
        \dot{\theta}_v &= -\frac{g_z}{L}\sin(\theta)-\frac{b}{m_p}\dot{\theta} 
    \end{align*}
    to simulate the motion of our gate, where $\theta$ is the angle displacement with respect to the vertical direction.
    We constrain the pendulum's motion in the $y-z$ plane, where $x=x_{f}$ and $v_x=0$.
    A Cartesian coordinate representation of the pendulum in the world frame $W$ can be
    obtained from the pendulum's angle displacement $\theta$ with respect to $\bs{P}_{WP}$ and $L$. 
    We can represent the state of the gate's center using the state vector 
    $\mb{x}_p=[x_p, y_p, z_p, v_{p, x}, v_{p, y}, v_{p, z}, q_{p, w}, q_{p, x}, q_{p, y}, q_{p, z}]$.
    
    \textbf{Model Predictive Control:}
    We solve the problem of passing through the swinging gate using non-linear model predictive control. 
    We make use of discrete time models, where a list of quadrotor states $\mb{x}_{q, h}, \forall h \in [0, H]$ and control 
    commands $\mb{u}_{q, h}, \forall h \in [0, H-1]$ are sampled with a discrete time step $d_t$.
    We define the objective $\mathcal{L}$ as a sum over three different cost components:
    a goal cost~$\mathcal{L}_\text{g}$, a tracking cost~$\mathcal{L}_\text{tr}$, and an action regularization cost~$\mathcal{L}_{\text{u}}$. 
    Thus, we solve the following constrained optimization problem:
    \begin{align*}
        \min_{\mathbf{u}_q, \mathbf{x}_q} & \quad \mathcal{L}_\text{g}(\mb{x}_{q, H}, \mb{r}_{g}) + \sum_{h=0}^{H-1} \mathcal{L}_\text{tr}(\mb{x}_{q, h}, \mb{r}_{h}, t_\text{tra}) + \mathcal{L}_{\text{u}}(\mb{u}_h) \\
         & = \bs{\delta}_{\text{g}, H}^T \mb{Q}_\text{g} \bs{\delta}_{\text{g}, H} + \sum_{h=0}^{H-1} \bs{\delta}_{\text{tr},h}^T \mb{Q}_\text{tr}(t_\text{tra}, h) \bs{\delta}_{\text{tr},h} + \bs{\delta}_{\text{u},h}^T \mb{Q}_{\text{u}} \bs{\delta}_{\text{u},h}\\
        \text{s.t.}: & \quad c_\text{min} \leq c \leq c_\text{max} \\
        & \quad  -\bs{\omega}_\text{max} \leq \bs{\omega}_B \leq \bs{\omega}_\text{max} 
    \end{align*}
    where $\bs{\delta}_{\text{tr}, h}= (\mb{x}_{\text{q}, h} - \mb{r}_{h})$ are differences between the vehicle's states $\mb{x}_{\text{q}, h}$
    and reference states $\mb{r}_{h}$ at the stage~$h$, and $\bs{\delta}_{\text{g}, H}= (\mb{x}_{\text{q}, H} - \mb{r}_{\text{g}})$ defines 
    the difference between the vehicle's terminal state~$\mb{x}_{\text{q}, H}$ and a hovering state~$\mb{r}_{\text{g}}$.
    Here, $\bs{\delta}_{u, h} = (\mb{u}_h - \mb{u}_{\text{r}})$ is a regularization for predicted 
    control commands~$\mb{u}_h$, where the reference command $\mb{u}_{\text{r}}=[g_z, 0, 0, 0]$ is the command 
    required for hovering the quadrotor.
    The control commands are constrained by $c_\text{min}, c_\text{max}, \bs{\omega}_\text{max} \in \mathbb{R}_{+}$.
    
    \textbf{Cost Functions:}
    In MPC, we minimize a sum of quadratic cost functions over the receding horizon $T$ using a sequential quadratic program~(SQP). 
    We design quadratic cost functions using positive definite diagonal matrices $\mb{Q}_\text{g}$, $\mb{Q}_\text{tr}(t_\text{tra}, h)$, and $\mb{Q}_\text{u}$.
    In particular, both $\mb{Q}_\text{g}$ and $\mb{Q}_\text{u}$ are time-invariant matrices.
    Here, $\mb{Q}_\text{g}$ defines the importance of reaching to a hovering state $\mb{r}_g$ at the end of the horizon and
    $\mb{Q}_\text{u}$ corresponds to the importance of taking the control commands that 
    are not diverging too much from the reference command~$\mb{u}_r$.
    
    Since the gate is swinging in the $y-z$ plane, in order to pass through the gate,
    the quadrotor has to fly forward in the $x$ direction and simultaneously minimize 
    its distance to the center of the gate in both $y$ and $z$ axes.
    Hence, the quadrotor has to track the pendulum's motion in both axes when it approaches to the gate. 
    To do so, we use a time-varying cost matrix $\mb{Q}_\text{tr}(\bs{\phi}, h)$, which is defined as:
    \begin{equation} \label{eq: exp_cost}
        \mb{Q}_\text{tr}(t_\text{tra}, h) = \mb{Q}_\text{tr, max} * \exp{ (- \alpha * (h * d_t - t_\text{tra})^2 ) }
    \end{equation}
    where the exponential function defines the temporal importance for each states $\mb{x}_{q}$,
    and $\alpha \in \mathbb{R}_{+}$ defines the temporal spread of states in terms of tracking the pendulum's motion.
    Here, $t_\text{tra} \in [0, 2T]$ is a time variable that defines the best traversal time for the quadrotor,
    having $ T < t_\text{tra} < 2T$ helps the quadrotor go to the hovering point after passing through the gate. 
    Hence, for states $\mb{x}_{q}$ that are close to the $t_\text{tra}$, we have $\mb{Q}_\text{tr}(t_\text{tra}, h) \approx (\mb{Q}_\text{tr, max}*1)$,
    which means that these states should strictly follow the pendulum in $y$ and $z$.
    However, for states that are faraway from $t_\text{tra}$, we have $\mb{Q}_\text{tr}(t_\text{tra}, h) \approx (\mb{Q}_\text{tr, max} * 0)$,
    which indicates that it is not necessary for these states to follow the pendulum's motion.
    Here, $\mb{Q}_\text{tr, max}$ defines the maximum weight that should be assigned for tracking the pendulum.
    Without considering the importance of each state at different time stages, e.g., weighting the
    tracking loss $\bs{\delta}_{\text{tr}, h}$ in all time stages using the same constant cost matrix, 
    the quadrotor flies trajectories that would oscillate around the forward axis~(see Fig.~\ref{fig: mpc_vs_upmpc}).
    
    Therefore, a key requirement for our MPC to solve the problem is to obtain
    the optimal traversal time $t_\text{tra}$ in advance. 
    A similar problem was discussed in~\cite{neunert2016fast}, where a time variable 
    at which a desired static waypoint should be reached by a quadrotor was determined by human experts. 
    In our case, the time variables are more difficult to obtain, especially when we consider adapting the
    variable online.
    
\subsection{Learning Traversal Time}
    We first consider the scenario where the quadrotor always starts from the same initial hovering state with
    $[x_q, y_q, z_q]=[-1, 0, 2]$ and the pendulum is hinged at a fixed pivot point $[x_f, y_f, z_f]=[2, 0, 3]$ 
    with cord length $L=2$ meter~(m). 
    The pendulum's initial angle and angular velocity are $[\theta, \dot{\theta}]= [\pi/2, 0]$ (in radians). 
    We define a hovering state $[x_g, y_g, z_g] = [4, 0, 2]$ as a goal state for the quadrotor to hover after passing through the gate.
    Given the dynamics of the vehicle and the pendulum, we want to plan a trajectory in the future time horizon
    $T=2$~seconds, such that the produced quadrotor trajectory $\bs{\tau}$ 
    intersects the center of the gate at the traversal time $t_\text{tra}$.

    We learn the decision variable $t_\text{tra}$ using Algorithm~\ref{algo: auto_mpc}~(Section \ref{section: method}), 
    where $t_\text{tra}$ is modeled as a high-level policy and is represented using a Gaussian distribution $t_\text{tra} \sim \pi(\mu, \sigma)$.
    We first sample a list of $t^{[i]}_\text{opt}, i\in [0\cdots N-1]$ of size $N$, and then,
    collect a vector of predicted trajectories $\bs{\tau}^{[i]}$ by solving $N$ MPC optimizations.
    We evaluate the sampled trajectories $\bs{\tau}$ using the following reward function:
    \begin{equation*}
        R(\bs{\tau}) = - \sum_{j} (\|\mb{x}_{q, j}-\mb{x}_{p, j}\|+\|\mb{y}_{q, j}-\mb{y}_{p, j}\|+\|\mb{z}_{q,j}-\mb{z}_{p, j}\|)
    \end{equation*}
    where $j \in [ (j^{\ast} -5) \cdots (j^{\ast}+4)]$ correspond to 10 time stages 
    that are close to the time stage determined by the samples $t_\text{tra}^{[i]}$ 
    via $j^{\ast} = \text{int}(t_\text{tra}/d_t)$.
    Maximizing this reward signal indicates that the high-level policy $\pi(\mu, \sigma)$ 
    tends to sample $t_\text{tra}$ that allows the MPC to plan a trajectory that has a minimum distance 
    between the quadrotor's state $\mb{x}_q$ and the center of the gate $\mb{x}_{p}$ during the traversal.
    This reward is maximized by solving the weighted maximum likelihood objective~(\ref{eq: wml}) using Algorithm~\ref{algo: auto_mpc}.

    \subsubsection{High-Level Policy Training}
\begin{figure}[!htp]
        \centering
        \includegraphics[width=0.45 \textwidth]{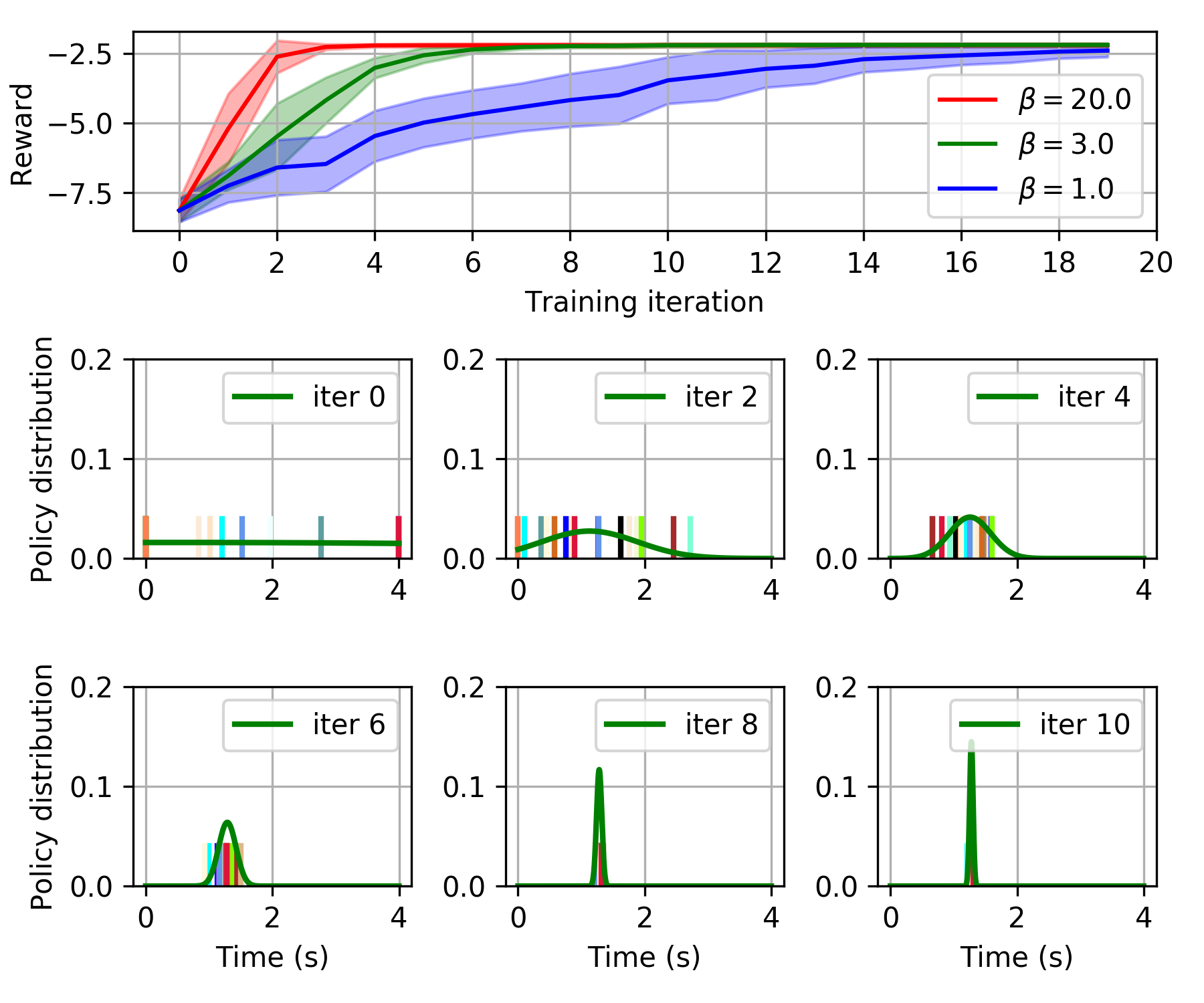}
        \caption{This figure shows the learning  progress of the high-level policy. 
        \textbf{Top: }Averaged rewards using different $\beta$ over 7 runs of each, where policies are randomly initialized with different random seeds.
        \textbf{Bottom: } A visualization of policy distributions and sampled $t_\text{tra}$ during training.
        The policy converges to an optimal solution after around 6 iterations.}
        \label{fig: learning_progress}
    \end{figure}

    Fig.~\ref{fig: learning_progress} shows the learning progress of the high-level policy. 
    The learning of such a high-level policy is extremely data-efficient and stable,
    where the policy converges in only a few trials. 
    For example, the policy is converged after around 6 training iterations when using $\beta=3.0$, 
    where in total $6\times30$ trajectories~(equivalent to 180 MPC optimizations) were sampled. 
    We use CasADi~\cite{casadi}, which is an open-source tool for nonlinear optimization and algorithmic differentiation,
    for our MPC implementation.
    We use a discretization time step of $dt = 0.04s$  and a prediction horizon of $T = 2s$.
    On average, each MPC optimization takes around $0.03 s$ on a standard laptop.

    \subsubsection{Traverse Trajectory Planning}
\begin{figure}[h]
        \centering
        \includegraphics[width=0.5 \textwidth]{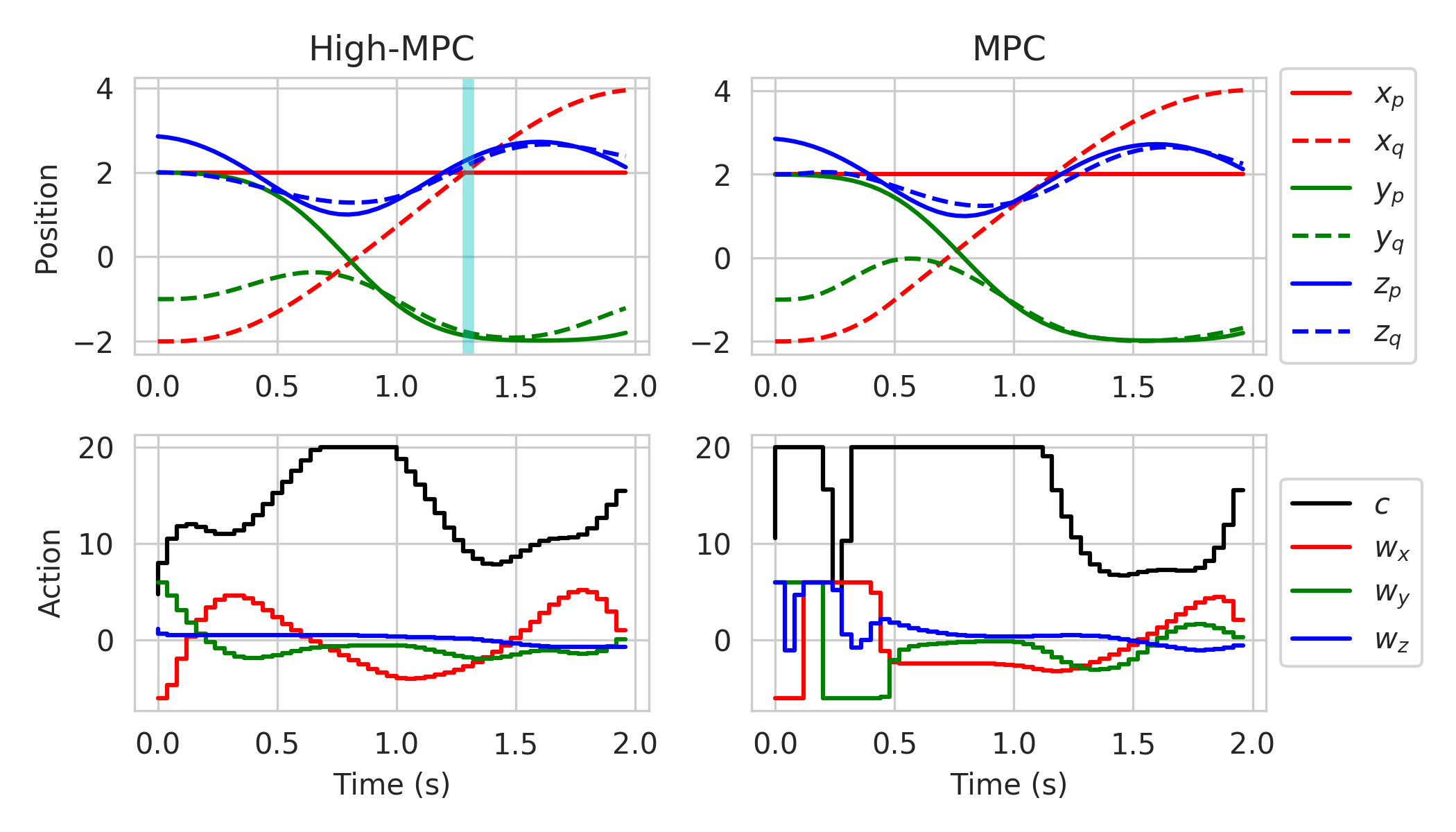}
        \caption{A comparison of planned trajectories between our High-MPC (with trained $t^{\ast}_\text{tra}=1.25$~(s)) 
        and a standard MPC. The vertical line indicates the passing moment.
        Our High-MPC is better for real-world deployment since the produced actions are
        much smoother than the standard MPC and reach the limit 
        for lower amount of time.}
        \label{fig: plan_vs}
    \end{figure}

    Fig.~\ref{fig: plan_vs} shows a comparison between the planned trajectory
    using our High-MPC (along with an optimized decision variable $t_\text{tra}=1.25$ seconds)
    and the solution from a standard MPC.
    The standard MPC minimizes the same cost function with a constant cost matrix $\mb{Q}_\text{tr}=\mb{Q}_\text{tr, max}$
    for all states and does not use the exponential weighting scheme. 
    As a result, both methods are capable of planning trajectories that pass through the swinging gate, 
    where absolute position errors at the traversal point in the $y-z$ plane are $0.24$ meters for High-MPC and $0.30$ meters for the standard MPC, respectively.
    Nevertheless, the control actions (the total thrust and body rates) produced by High-MPC 
    are better for real-world deployment since the inputs reach their limit for lower amount of time, 
    leaving more control authority to counteract disturbance.
    Our approach only tries to follow the pendulum's motion in $y$ and $z$ directions at
    the time stages closed to the learned traversal time $t_\text{tra}$.

\subsection{Learning Adaptive Traversal Time}
    \begin{figure*}[t]
        \centering
        \begin{tabular}{@{}c@{}}
           \includegraphics[width=0.92\linewidth]{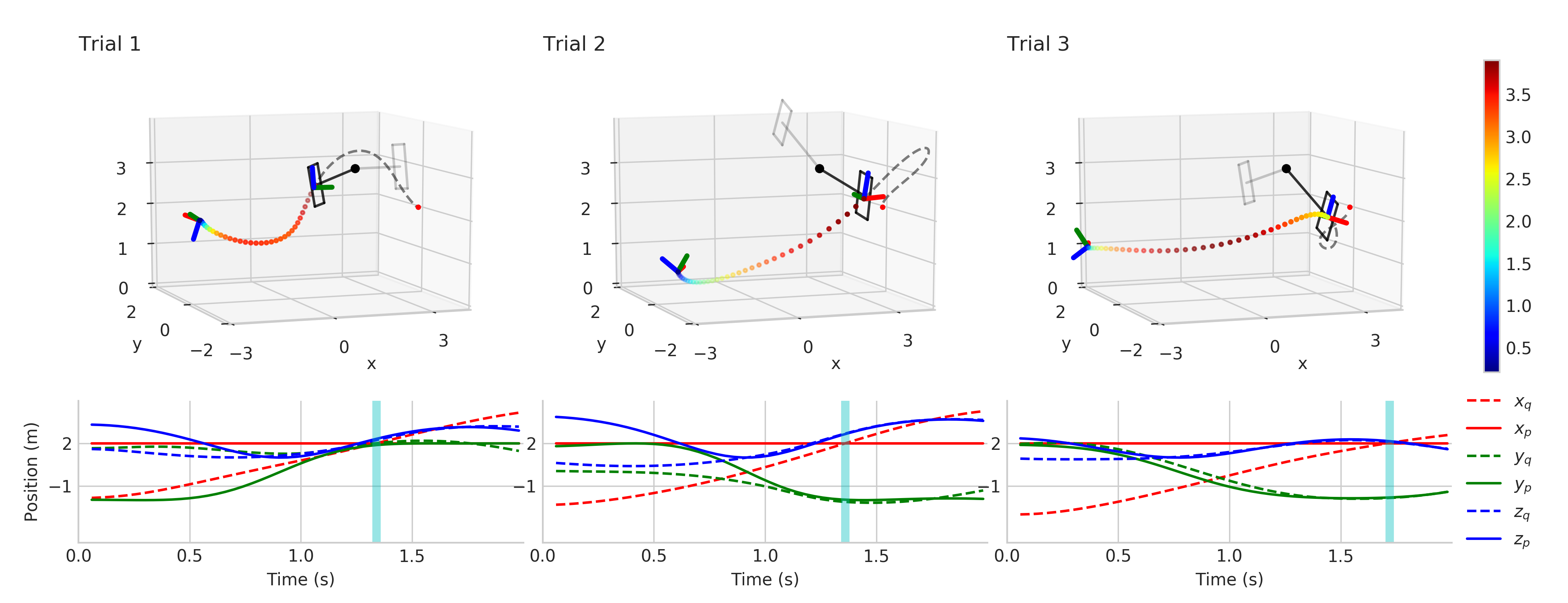}
        \end{tabular}
        \begin{tabular}{@{}c@{}}
           \includegraphics[width=0.92\linewidth]{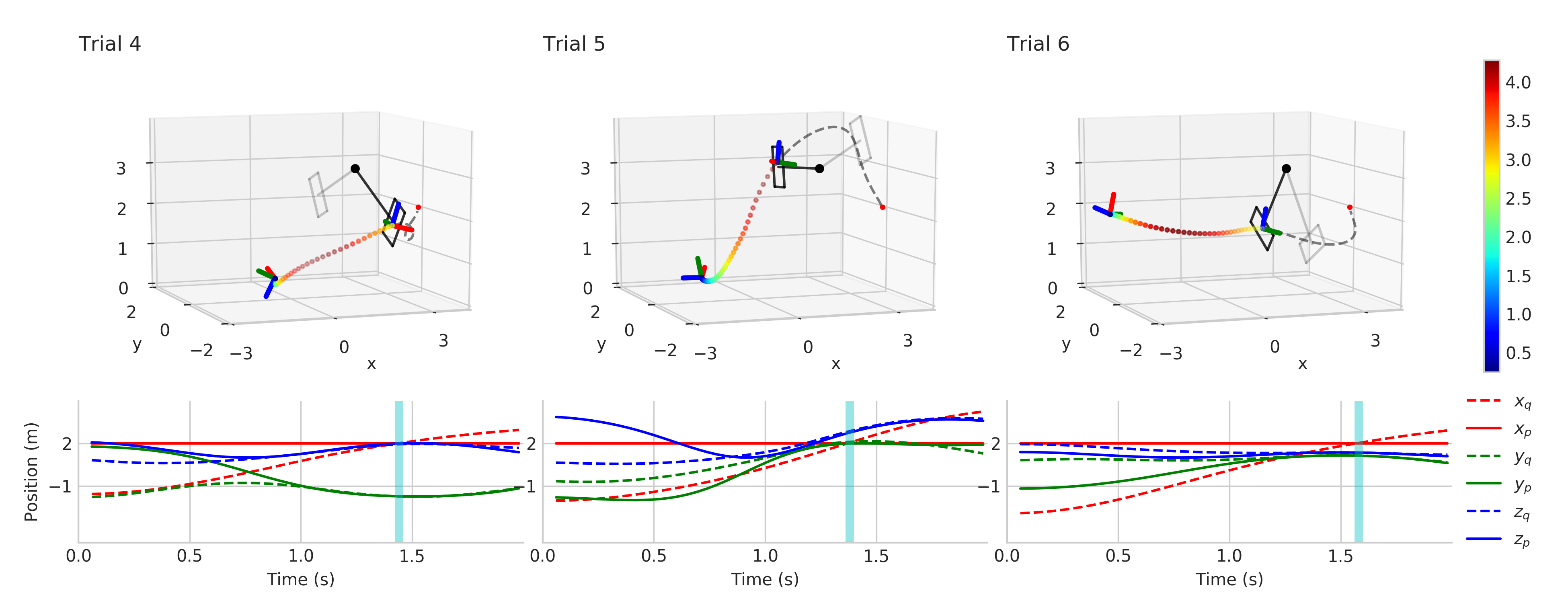}
        \end{tabular}
    \caption{Demonstrations of our High-MPC for flying through a swinging gate. 
    The initial states of the quadrotor and the pendulum are randomly initialized. 
    In the 3D plots, the initial states of the pendulum are indicated by the grey color, and the black gates
    show the moment when the quadrotor is intersecting in the gate. 
    The color bars on the right side specify the quadrotor speed in the $x$ direction.
    The grey dash lines are planned trajectory by our MPC and colored dots are traveled trajectories.
    The quadrotor's body frame is indicated by $[ {\color{red} x_q}, {\color{green} y_q}, {\color{blue} z_q}]$. 
    The 2D plots show travelled trajectories of the quadrotor and the pendulum.}
    \label{fig: upmpc_examples}
    \end{figure*}
    
    Learning a single high-level policy without taking the robot's observation into account 
    is only useful for selecting time-invariant variables or for planning a one-shot trajectory, 
    where the dynamics are perfectly modeled.
    This, however, is generally not the case.
    For example, our task requires the MPC to constantly update its prediction based on the
    the vehicle's state with respect to that of the dynamic gate. 
    Hence, we also want to find a high-level policy which is capable of adaptively selecting the
    time variable $t^{\ast}_\text{tra}$ depending on the robot's observation.
    
    \subsubsection{Deep High-Level Policy Training}
    To do so, we make use of a multilayer perceptron~(MLP) to generalize the $t^{\ast}_\text{tra}$ to
    different contexts~$\mb{o}_t$.
    We represent $\mb{o}_t$ as an observation of the vehicle using~$\mb{o}_t=\mb{x}_{q, t}- \mb{x}_{p, t}$, which represents 
    the difference between the vehicle's state $\mb{x}_{q, t}$ and the pendulum's state~$\mb{x}_{p, t}$ at time step $t$.
    We use Algorithm~\ref{algo: online_mpc}~(Section \ref{section: method}), where we combine the learning of an optimal 
    high-level policy online with a supervised learning approach to train the MLP.
    We first randomly initialize the system, which means we use random initial states for the
    quadrotor, and drop the pendulum from random angles; then, we find the optimal traversal time $t^{\ast}_\text{tra}$ at this state. 
    We solve the MPC optimization using $t^{\ast}_\text{tra}$ and apply the optimal control action to a simulated quadrotor.
    We repeat this process again at each simulation time step until the quadrotor flies through the gate or it reaches the maximum simulation steps. 
    In total, we collect 40,000 samples that consist of observation-traversal-time pairs $(\mb{o}_t, t_\text{tra})$.
    It takes a single core CPU several hours to collect the data, however, the total sampling time can be significantly reduced
    using parallel processing or multithreading.
    We use Tensorflow~\cite{tensorflow2015-whitepaper} to implement the a fully-connected MLP with two hidden layers of 32 units, 
    and \textit{ReLU} nonlinearities. The training of network weights takes less than 5 minutes on a notebook with
    a \textit{Nvidia Quadro P1000} graphics card. 

    \begin{figure}[!htp]
        \centering
        \includegraphics[width=0.45 \textwidth]{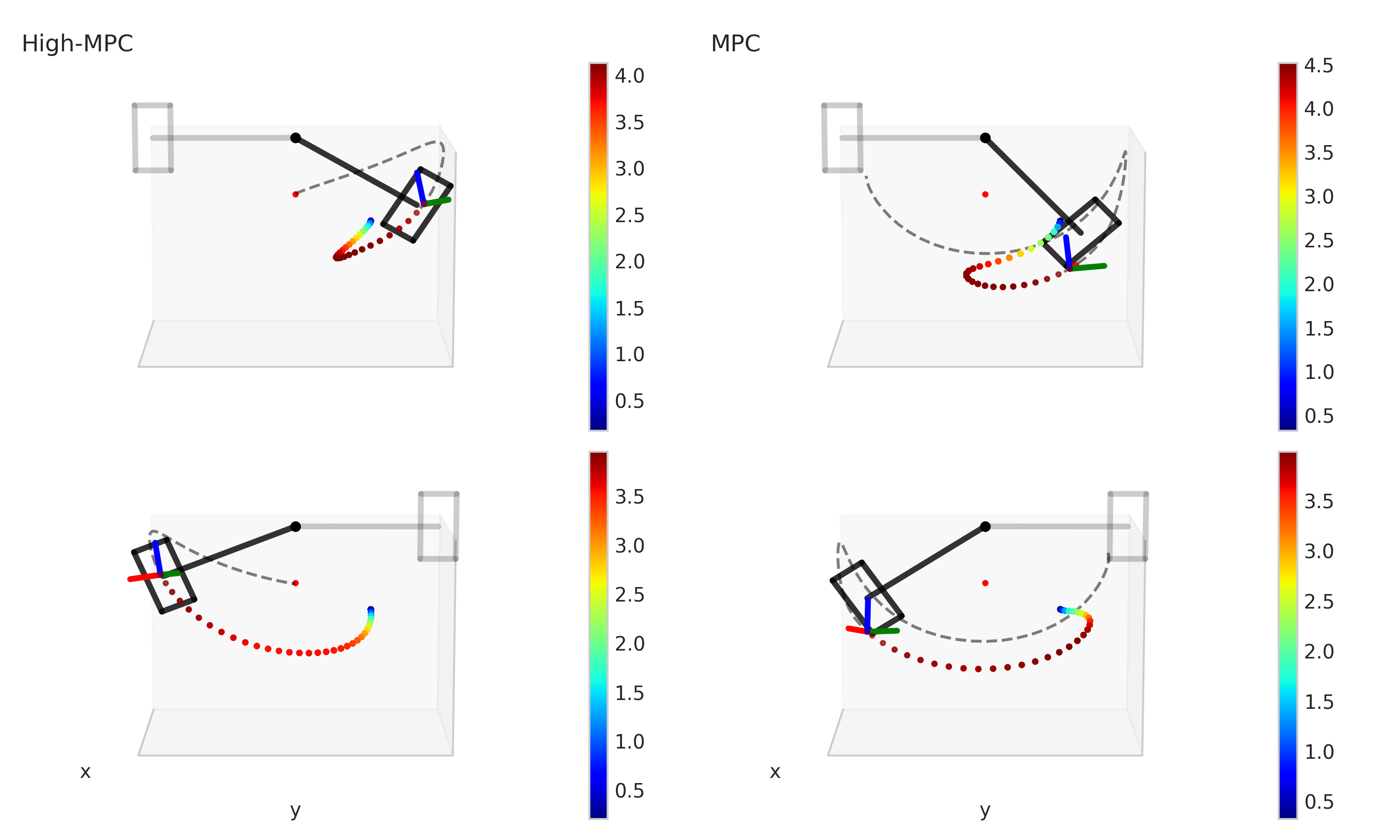}
        \caption{Comparisons between our High-MPC~(left) and a standard MPC~(right),
        where initial states of the system are the same for both methods.
        \textbf{Top: } the swinging gate is released from $\theta=1.57$~(rad).
        \textbf{Bottom: } the swinging gate is released from $\theta=-1.57$~(rad).}
        \label{fig: mpc_vs_upmpc}
    \end{figure}
    
    \subsubsection{Passing Through a Fast Moving Gate via High-MPC}
    We evaluate the effectiveness of our High-MPC by controlling a simulated quadrotor to pass through a fast swing gate,
    where the quadrotor and the pendulum are randomly initialized in different states. 
    Based on the state of the quadrotor, the motion of the pendulum (including 2s of predicted pendulum motion in the future), 
    and the predicted traversal time, our High-MPC simultaneously plans a trajectory and controls the vehicle to pass through the gate.
    Fig.~\ref{fig: upmpc_examples} shows six random examples of the quadrotor successfully flying through 
    the swinging gate.

    In addition, we compared the performance of our High-MPC to a standard MPC~(Fig.~\ref{fig: mpc_vs_upmpc}), where
    the standard MPC optimizes a cost function without considering the temporal importance 
    of difference states in tracking the pendulum motion.
    The standard MPC failed to pass through the gate and results in trajectories that are oscillating about
    the forward direction~($x$ axis).

\section{Discussion and conclusion}

\label{section: conclusion}
In this work, we introduced the idea of formulating the design 
of hard-to-engineer decision variables in MPC as a probabilistic inference problem, 
which can be solved efficiently using an EM-based policy search algorithm.  
We combined self-supervised learning with the policy search method to 
train a high-level neural network policy.
After training, the policy is capable of adaptively making online decisions for the MPC. 
We demonstrated the success of our approach by combining
a trained MLP policy with a MPC to solve a 
challenging control problem, where the task is to maneuver a quadrotor 
to fly through the center of a fast-moving gate.
We compared our approach~(High-MPC) to a standard MPC and showed that ours 
achieve more robust results, and hence, it is more promising to deploy our method 
on real robots, thanks to the online decision variable adaptation 
scheme realized by the deep high-level policy.
Besides, our approach has the advantage of tightly coupling planning
and optimal control together, and hence, forgo the need for 
an explicit trajectory planner.

Nevertheless, our approach has limitations such as it 
requires multiple MPC optimizations in-the-training-loop in order to find optimal variables.
It is possible to learn a vector of high-dimensional decision variables and more complex neural network policies, 
however, the sample complexity will increase by a large margin. 
To fully exploit the potential of automatically learning high-level policies for optimal control,
we hope that our work sparks more researchers' interests in this domain to derive new algorithms
and opens up opportunities for solving more complex robotic problems, such as 
real-world robot navigation in a complex dynamic environment.
To test the scalability and generalization of our High-MPC, in the near future we intend to deploy the algorithm on a real robot system.




\bibliographystyle{IEEEtran}
\bibliography{references}

\end{document}